\documentclass[runningheads]{llncs}
\usepackage{graphicx}
\usepackage{amssymb}
\usepackage{amsmath}
\usepackage{multicol}
\usepackage{bigstrut}

\begin{document}
\title{A gray-box approach for curriculum learning}
\author{Francesco Foglino\inst{1} \and
Matteo Leonetti\inst{1}\orcidID{0000-0002-3831-2400} \and\\
Simone Sagratella\inst{2}\orcidID{0000-0001-5888-1953} \and\\
Ruggiero Seccia\inst{2}\orcidID{0000-0001-5292-1774} }
\authorrunning{F. Foglino et al.}
\institute{School of Computing, University of Leeds, Leeds, UK\\
\email{\{scff,M.Leonetti\}@leeds.ac.uk} \and
Department of Computer, Control and Management Engineering Antonio Ruberti, Sapienza
University of Rome, Via Ariosto 25, 00185 Roma, Italy
\email{\{sagratella,seccia\}@diag.uniroma1.it}}
\maketitle              % typeset the header of the contribution
\begin{abstract}
Curriculum learning is often employed in deep reinforcement learning to let the agent progress more quickly towards better behaviors.
Numerical methods for curriculum learning in the literature provides only initial heuristic solutions, with little to no guarantee on their quality.
We define a new gray-box function that, including a suitable
scheduling problem, can be effectively used to reformulate the curriculum learning problem.
We propose different efficient numerical methods to address this gray-box reformulation.
Preliminary numerical results on a benchmark task in the
curriculum learning literature show the viability of the proposed approach.

\end{abstract}
\section{Introduction}

Curriculum learning is gaining popularity in (deep) reinforcement
learning, see e.g. \cite{curriculum2019preprint} and references therein.
It can provide
better exploration policies through transfer and generalization from
less complex tasks.
Specifically, curriculum learning is often employed to
let the agent progress more quickly towards better behaviors, thus having the
potential to greatly increase the quality of the behavior discovered
by the agent. However, at the moment, creating an appropriate
curriculum requires significant human intuition, e.g.
curricula are mostly designed by hand. 
Moreover, current methods for automatic task sequencing
for curriculum learning in reinforcement learning provided only initial
heuristic solutions, with little to no guarantee on their quality.

After a brief introduction to reinforcement learning, see e.g. \cite{leonetti2012combining,mnih2015human},
we define the curriculum learning problem.
This is an optimization problem that cannot be solved with standard methods for nonlinear
programming or with derivative-free algorithms.
We define a new gray-box function that, including a suitable scheduling problem, can be effectively used to reformulate the curriculum learning problem. This gray-box reformulation can be addressed in different ways.
We investigate both heuristics to estimate approximate solutions of the gray-box problem, and
derivative-free algorithms to optimize it.
Finaly, preliminary numerical results on a benchmark task in the
curriculum learning literature show that the proposed gray-box methods can be efficiently used to
address the curriculum learning problem.

\section{Reinforcement learning background}

Consider an {\bf agent} that acts in an environment $m$ according to a {\bf policy} $\pi$.
The policy $\pi$ is a function that given a state $s \in S$ and a possible action $a \in A$ returns a number in $[0,1]$ representing the probability that the agent executes the action $a$ in state $s$.
The {\bf environment} $m$ is modeled as an episodic Markov Decision Process (MDP), that is, a tuple
$(S,A,p_m,r_m,T_m,\gamma_m)$, where both $S \subset \mathbb{R}^d$ and $A \subset \mathbb{R}^d$ are nonempty and finite sets, for which $d > 0$ is the state dimension, $p_m : S \times A \to S$ is a transition function, $r_m : S \times A \to \mathbb R$ is a reward function, $T_m \in \mathbb N$ is the maximum length of an episode, and $\gamma_m \in [0,1]$ is a parameter used to discount the rewards during each episode. At every time step $t$, the agent perceives the state $s^t$, chooses an action $a^t$ according to $\pi$, and the environment transitions to state  $s^{t+1} = p_m(s^t, a^t)$. We assume for simplicity for the transition function $p_m(s^t, a^t) \triangleq P_S(s^t + a^t)$, where $P_S$ denotes the projection operator over the set $S$. Every episode starts at state $s^0 \in S$, where $s^0$ can depend both on the environment $m$ and the episode. During an episode, the agent receives a total reward of 
$$
\displaystyle R_m(s^0,\ldots,s^{T_m-1},a^0,\ldots,a^{T_m-1}) \triangleq \sum_{t=0}^{T_m-1} (\gamma_m)^t r_m(s^t,a^t).
$$
Note that $\gamma_m$ is used to emphasize the rewards that occur early during an episode. We say that $s^{\overline t}$ is an absorbing state if $s^t = s^{\overline t}$ for all $t \geq \overline t$, and $r_m(s^{\overline t}, a) = 0$ for any action $a \in A$, that is, the state can never be left, and from that point on the agent receives a reward of $0$. Absorbing states effectively terminate an episode before the maximum number of time steps $T_m$ is reached. 

The policy function $\pi$ is obtained from an estimate $\widehat q_{\pi}$ of the value function
$$
q_{\pi}(s,a) \triangleq \mathbb E \left[\sum_{j=t}^{T_m-1} (\gamma_m)^{j-t} r_m(s^j,a^j) \, : \, s^{t} = s, a^{t} = a\right],
$$
for any state $s \in S$ and action $a \in A$.
The value function is the expected reward for taking action $a$ in state $s$ at any possible time step $t$ and following $\pi$ thereafter until the end of the episode.
We linearly approximate the value function $q_{\pi}$ in a parameter $\theta \in D \subset \mathbb R^K$:
$$
\widehat q_{\pi} (s,a;\theta) \triangleq \sum_{k=1}^K \theta_k \phi_k (s,a),
$$
where $\phi_k$ are suitable basis functions mapping the pair $(s,a)$ into $\mathbb R$.
The policy function $\pi$ for any point $(s,a) \in S \times A$ and any parameter $\theta \in D$, is given by
$$
\displaystyle \pi(s,a;\theta) \triangleq \frac{\widehat q_{\pi} (s,a;\theta)}{\displaystyle \sum_{\alpha \in A} \widehat q_{\pi} (s,\alpha;\theta)}.
$$

\noindent
During the {\bf reinforcement learning process} the policy $\pi$ is optimized by varying the parameter $\theta$ over $D$ in order to obtain greater values of the environment specific total reward $R_m$.
In this respect, we introduce the black-box function $\psi_m : \mathbb R^K \to \mathbb R$, which takes the parameter $\theta$ and returns the expected total reward $\mathbb E[R_m]$  obtained with the policy $\pi(\, \cdot \,, \,\cdot \,;\theta)$.
It is reasonable to assume that $\psi_m$ is bounded from above over $D$.
Then, a global optimal policy for the environment $m$ is given by $\widehat \theta \in D$ satisfying
\begin{equation}\label{eq: optimum}
\psi_m(\widehat \theta) \geq \psi_m(\theta), \qquad \forall \, \theta \in D.
\end{equation}
In practical reinforcement learning optimization, at any time step $t$ of a finite number $N_m$
of episodes the policy parameter $\theta$ is updated by using a learning algorithm that exploits the value of both the reward $r_m(s^t,a^t)$ and the current estimate of the value function $\widehat q_\pi$, aiming at computing a point $\widehat \theta$ satisfying (\ref{eq: optimum}).
Certainly, the better the point $\overline \theta \in D$ from which the learning procedure starts, the faster the global optimum is achieved.
In general, due to the limited number $T_m N_m$ of iterations granted, we say that the learning algorithm is able to compute a local optimal $\widehat \theta \in D$ satisfying
\begin{equation}\label{eq: local optimum}
\psi_m(\widehat \theta) \geq \psi_m(\theta), \qquad \forall \, \theta \in D \, : \, \|\theta - \overline \theta\| < \zeta_m,
\end{equation}
where $\zeta_m > 0$ is related to $T_m N_m$ and $\overline \theta \in D$ is the starting guess.

\section{The curriculum learning problem}

We want the agent to quickly obtain great values of $\psi_{m_L}$ in a specific environment $m_{L}$ that we call the {\bf final task}.
To do this, it is crucial to ensure that the reinforcement learning phase in the final task $m_{L}$ starts from a good initial point $\theta^{L}$ ideally close to a global maximum of $\psi_{m_L}$ over $D$.
Curriculum learning is actually a way to obtain a good starting point $\theta^{L}$ computed by sequentially learning the policy on a subset of possible tasks (i.e. environments) different from the final task $m_L$, see e.g. \cite{curriculum2019preprint} and references therein.
The {\bf curriculum} $c = (m_0, \ldots,m_{L-1})$ is the sequence of these tasks in which the policy of the agent is optimized before addressing the final task $m_L$.
Specifically, given a starting $\theta^0 \in D$, the point $\theta^1$ is obtained by (approximately) maximizing $\psi_{m_0}$ over $\{\theta \in D \, : \, \|\theta - \theta^0\| < \zeta_{m_0} \}$, the point $\theta^2$ is obtained by (approximately) maximizing $\psi_{m_1}$ over $\{\theta \in D \, : \, \|\theta - \theta^1\| < \zeta_{m_1} \}$, and so on. At the end of this process we get a point $\theta^{L}$ ready to be used as starting guess for the optimization of the policy in the final task $m_L$. Clearly, the obtained $\theta^{L}$ depends on the specific sequence of tasks in the curriculum $c$. To underline this dependence, we write $\theta^{L}(c)$.

We denote with $\mathcal T$ the set of $n$ available tasks. The tasks in $\mathcal T$ must be included in the curriculum $c$ of length less than $L \leq n$ in a specific order and without repetitions. The quality of the curriculum $c$ is given by $\psi_{m_L}(\theta^{L+1}(c))$ that is obtained by executing learning updates with respect to $\psi_{m_L}$ for a finite number $N_{m_L}$ of episodes and starting from $\theta^{L}(c)$.
A practical performance metric of great interest is given by the so called {\bf regret function}, which takes into account both the expected total reward that is obtained for the final task at the end of the learning process, and how fast it is achieved:
$$
\displaystyle {\cal P}_r (c) \triangleq \sum_{i=1}^{N_{m_L}} \left( g - \psi_{m_L} \left(\theta^{L+(i / N_{m_L})}(c)\right) \right),
$$
where $g$ is a given good performance threshold (which can be the total reward obtained with the optimal policy when known), and $\theta^{L+(i / N_{m_L})}(c)$ is the point obtained with the learning algorithm at the end of the $i$th episode. Given the curriculum $c$, the function ${\cal P}_r (c)$ sums the gaps between the threshold $g$ and the total reward actually achieved at every episode. Clearly the aim is to minimize it
\begin{equation}\label{eq: min regret}
  \underset{c \in {\cal C}}{\text{minimize }} \quad {\cal P}_r (c),
\end{equation}
where ${\cal C}$ is the set of all feasible curricula obtained from ${\cal T}$.

Problem (\ref{eq: min regret}) presents two main drawbacks:
(i) having a black-box nature, its objective function has not an explicit definition and it is in general nonsmooth, nonconvex, and even discontinuous;
(ii) it is a constrained optimization problem, whose feasible set is combinatorial.
With the aim of solving problem (\ref{eq: min regret}), drawback (i) does not allow us to resort to methods for general Mixed-Integer NonLinear Programs (MINLP), see e.g. \cite{belotti2013mixed}, while (ii) makes it difficult to use standard Derivative-Free (DF) methods, see e.g. \cite{custodio2017methodologies,di2016direct}.
See \cite{curriculum2019preprint} and the references therein for possible numerical procedures to tackle problem (\ref{eq: min regret}).
As we show in section \ref{sec: algorithms}, the methods proposed in \cite{curriculum2019preprint} constitute only a preliminary step in order to solve efficiently the curriculum learning problem.

In the next section we define a new gray-box reformulation for problem (\ref{eq: min regret}) that incorporates a scheduling problem. Afterwards, we propose different practical techniques to address this gray-box reformulation.

\section{The scheduling problem to minimize regret}

Let us introduce the variables $\delta \in \{0,1\}^n$ and $\gamma \in \{0,1\}^{n \times (n-1)}$.
Any $\delta_i$ indicates the presence of the $i$th task of $\cal T$ in the curriculum $c$, specifically, $\delta_i = 1$ if and only if the $i$th task of $\cal T$ is in the curriculum $c$.
Any $\gamma_{ij}$, with $i \neq j$, is an indicator variable used to model the order of the task in the curriculum: $\gamma_{ij} = 1$ if and only if the $i$th task of $\cal T$ is in the curriculum $c$ and it is scheduled before the $j$th task of $\cal T$. All the tasks not included in the curriculum are considered scheduled after all the ones included.

Minimizing the regret ${\cal P}_r$ is equivalent to maximizing the merit function $U$ given by 
$$
U(c) \triangleq \sum_{i=1}^{N_{m_L}} \psi_{m_L} \left(\theta^{L+(i / N_{m_L})}(c)\right).
$$
We make the following assumption:
\begin{description}
 \item[(A1)]
Every task $m_i$ in $c$ contributes to the value of $U$ with a fixed individual utility $u_i \geq 0$. Moreover, considering all pairs $(i,j) \in \{1,\ldots,n\} \times \{1,\ldots,n\}$ with $i \neq j$, if the $i$th task of $\cal T$ is in the curriculum $c$ and it is scheduled before the $j$th task of $\cal T$, then there is a penalty in $U$ equal to $p_{ij} \geq 0$.
\end{description}
This concept of penalty in assumption (A1) is useful to model the fact that a task $m_j$ can be preparatory for another task $m_i$. In this sense, if the policy is not optimized in the preparatory task $m_j$ before it is optimized in task $m_i$, then the utility given by task $m_i$ has to be reduced by the corresponding penalty.

We intend to approximate $U$ with the following function that is linear with respect to $(\delta, \gamma)$:
$$
\widehat U(\delta,\gamma;u,p) \triangleq \sum_{i = 1}^n u_i \delta_i - \sum_{i=1}^n \sum_{i \neq j=1}^n p_{ij} \gamma_{ij}.
$$
If assumption (A1) holds, then certainly $\widehat U$ is a good approximation of $U$. In general cases, given the utilities $u$ and the penalties $p$, our idea is to maximize $\widehat U$ by modifying the indicator variables $\delta$ and $\gamma$ corresponding to feasible curricula in $\cal C$.
We introduce additional variables $x \in [0,L-1]^n \cap \mathbb Z^n$ indicating the order of the tasks in the curriculum $c$; if the $i$th task of $\cal T$ is not in $c$ then $x_i = L-1$.
We are ready to define the scheduling problem for curriculum learning.
\begin{align}\label{eq: scheduling}
  \underset{x, \delta, \gamma}{\text{maximize }} \quad & \widehat U(\delta,\gamma;u,p) \nonumber\\[5pt]
\text{subject to } \quad & x_i \geq (L-1) (1-\delta_i), \quad i = 1, \ldots, n \nonumber\\[5pt]
                         & x_i + \delta_j \leq x_j + L\gamma_{ji}, \quad i = 1, \ldots, n, \; j = 1, \ldots, n, \; i \neq j \\[5pt]
                         & \gamma_{ij} + \gamma_{ji} \leq 1, \quad i = 1, \ldots, n, \; j = 1, \ldots, n, \; i \neq j \nonumber\\[5pt]
                         & x \in [0,L-1]^n \cap \mathbb Z^n, \quad \delta \in \{0,1\}^n, \quad \gamma \in \{0,1\}^{n \times (n-1)}. \nonumber
\end{align}
Problem (\ref{eq: scheduling}) is an Integer Linear Program (ILP) that can be solved by resorting to many algorithms in the literature.

The following properties hold:
\begin{itemize}
\item Let $(\widehat x, \widehat \delta, \widehat \gamma)$ be an optimal point of the scheduling problem (\ref{eq: scheduling}) with $(u, p) \in \mathbb R^n_+ \times \mathbb R^{n \times (n-1)}_+$.
Let $\widehat c = (\widehat m_0, \ldots, \widehat m_{L-1})$ be such that, for all $j \in \{0, \ldots, L-1\}$, $\widehat m_j = m \in {\cal T}$ with $\widehat x_{\text{ord}(m)} = j$ and $\widehat \delta_{\text{ord}(m)} = 1$, where the operator ord($m$) returns the index of the task $m$ in $\cal T$. Then $\widehat c \in {\cal C}$, i.e. $\widehat c$ is a feasible curriculum.
\item Let $\widehat c = (\widehat m_0, \ldots, \widehat m_{L-1})$ be any curriculum in ${\cal C}$, then parameters $(\widehat u, \widehat p) \in \mathbb R^n_+ \times \mathbb R^{n \times (n-1)}_+$ exist such that solving problem (\ref{eq: scheduling}) with $(u, p) = (\widehat u, \widehat p)$ gives $\widehat x$ such that $\widehat x_{\text{ord}(\widehat m_j)} = j$ and $\widehat \delta_{\text{ord}(\widehat m_j)} = 1$, for all $j \in \{0, \ldots, L-1\}$. That is, any curriculum in $\cal C$ can be computed by solving problem (\ref{eq: scheduling}) with suitable parameters $(u, p)$.
\end{itemize}

\noindent
We introduce the gray-box function $\Psi : \mathbb R^{n \times n} \to \mathbb R$, which takes the parameters $(u, p)$, computes a curriculum $c$ by solving problem (\ref{eq: scheduling}) with parameters $(u, p)$, and returns the regret ${\cal P}_r (c)$.
By using the gray-box  function $\Psi$, problem (\ref{eq: min regret}) can be equivalently reformulated as
\begin{equation}\label{eq: gray-box problem}
  \underset{(u,p) \in \mathbb R^n_+ \times \mathbb R^{n \times (n-1)}_+}{\text{minimize }} \quad \Psi(u,p).
\end{equation}

\section{Numerical methods for the gray-box}

The gray-box function $\Psi$ can be used in different ways in order to solve the curriculum learning problem efficiently.
Here we consider three of them.

\begin{itemize}
 \item Problem (\ref{eq: gray-box problem}) is a black-box optimization problem whose feasible set includes only lower bounds. Therefore we can resort to many DF algorithms in order to compute (approximate) optimal points of (\ref{eq: gray-box problem}). A potential solution is represented by Sequential Model-Based Optimization (SMBO) methods which consider the information obtained by all the previous iterations to build a surrogate probabilistic model of $\Psi(u,p)$. At each iteration a new point is drawn by maximizing an acquisition function and the information gained with this new sample is used to update the surrogate model \cite{frazier2018tutorial,Larochelle,Outoftheloop}.

 \item We can compute a good estimate for $(\overline u,\overline p)$ and then evaluate $\Psi(\overline u,\overline p)$ in order to have a good value of the regret.
 
 \item We can use a good estimate for $(\overline u,\overline p)$ as a reference point to define a trust region for the feasible set of problem (\ref{eq: gray-box problem}). The resulting furtherly constrained black-box optimization problem can be solved with many DF algorithms such as a Tree-structured Parzen Estimator (TPE), see e.g. \cite{bengio}, which allows us to define a distribution of probability of the parameters ($u,p$) to optimize.
\end{itemize}

\noindent
Computing a good estimate for $(\overline u,\overline p)$ can be critical for obtaining good numerical performances.
Here we propose a method that is justified by the assumption (A1). In that, if the assumption (A1) holds, then we have for any $(i,j)$ with $i \neq j$:
\begin{align*}
&U(m_i,m_j) = \overline u_i + \overline u_j - \sum_{k=1, \, k\neq i}^n \overline p_{ik} - \sum_{k=1, \, j\neq k\neq i}^n \overline p_{jk} + \overline U, \\
&U(m_i) = \overline u_i - \sum_{k=1, \, k\neq i}^n \overline p_{ik} + \overline U, \qquad U(m_j) = \overline u_j - \sum_{k=1, \, k\neq j}^n \overline p_{jk} + \overline U,
\end{align*}
where $\overline U$ is an unknown constant.
That implies
\begin{align}
 \overline p_{ji} &= U(m_i,m_j) - U(m_i) - U(m_j) + \overline U, \label{eq: p}\\
 \overline u_i &= U(m_i) + \sum_{k=1, \, k\neq i}^n \overline p_{ik} - \overline U \nonumber\\
                &= U(m_i) + \sum_{k=1, \, k\neq i}^n \left( U(m_k,m_i) - U(m_k) - U(m_i) \right) + (n-2) \overline U. \label{eq: u}
\end{align}
We observe that computing this estimate requires $n^2$ evaluations of $U$.

In the following section we adapt these ideas to a benchmark task in the curriculum learning literature.

\section{Experimental evaluation}\label{sec: algorithms}

In order to evaluate the effectiveness of the proposed framework, we implemented it on the GridWorld domain. In this section, we describe the GridWorld's setting and all the libraries adopted for the definition of the framework.

\subsection{GridWorld}

\begin{figure}
\centering
\includegraphics[width=.25\textwidth]{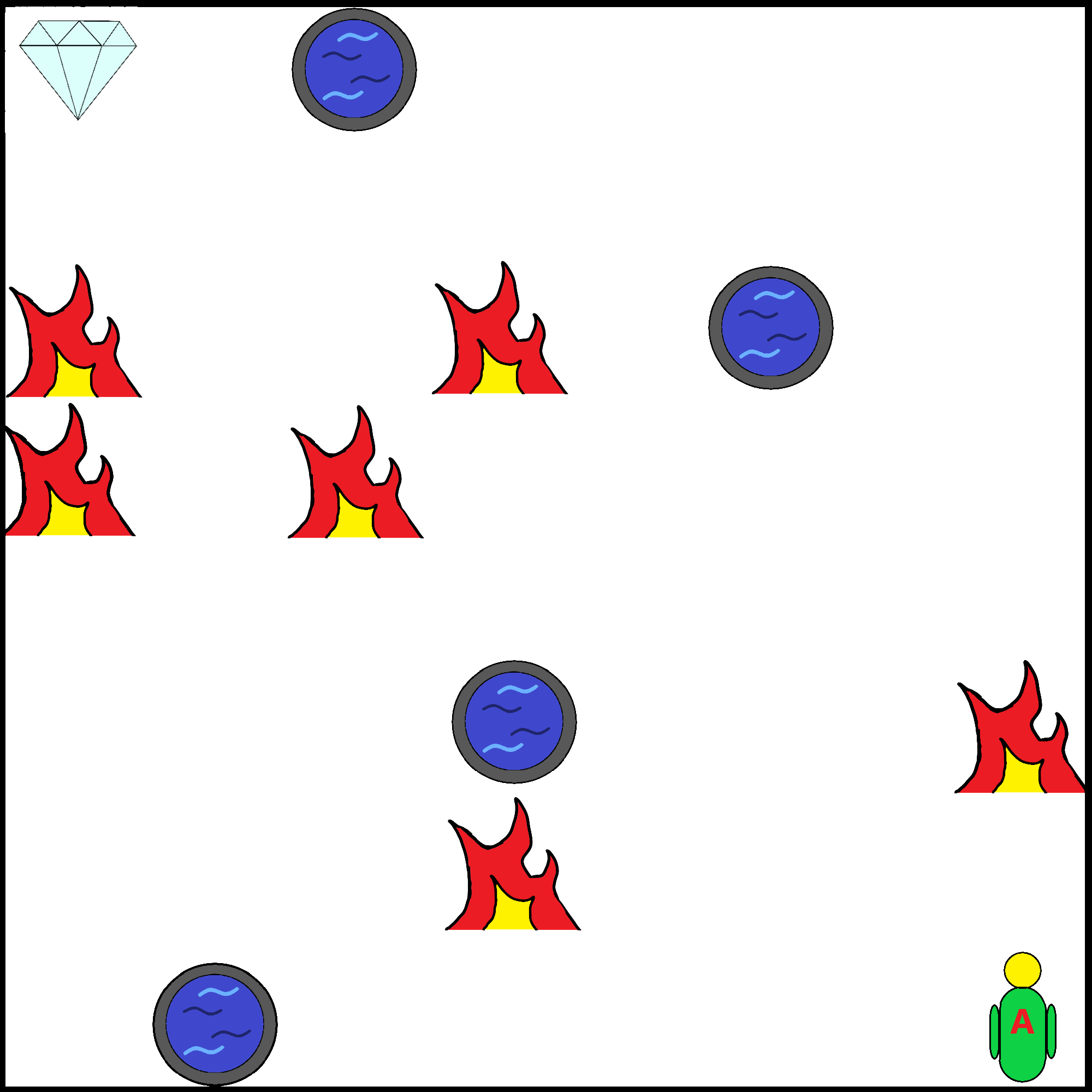}
\caption{An example of GridWorld.} \label{fig1}
\end{figure}

GridWorld is an implementation of an episodic grid-world domain
used in the evaluation of existing curriculum learning methods, see e.g. \cite{svetlik2017automatic}.
Each cell can be free, or occupied by a fire, pit, or treasure. The aim of the game is to find the treasure in the least number of possible episodes, avoiding both fires and pits. An example of GridWorld is shown in Figure \ref{fig1}.
\begin{description}
 \item[States $S$: ] The state is given by the agent position, that is $d = 2$.
 \item[Actions $A$ and transition function $p_m$: ] The agent can move in the four cardinal directions, and the actions are deterministic.
 \item[Reward function $r_m$: ] The reward is -2500 for entering a pit, -500 for entering a fire,
-250 for entering the cell next to a fire, and 200 for entering a
cell with the treasure. The reward is -1 in all other cases.
 \item[Episodes length $T_m$, absorbing states, discount parameter $\gamma_m$: ] All the \\ episodes terminate under one of these three conditions: the agent
falls into a pit, reaches the treasure, or executes a maximum number of actions ($T_m = 50$). We use $\gamma_m = 0.99$.
 \item[Basis functions $\phi_k$: ]   The variables fed to tile coding are the distance from, and relative position of,
the treasure (which is global and fulfills the Markov property), and
distance from, and relative position of, any pit or fire within a radius of 2 cells from the
agent (which are local variables, and allow the agent to learn how
to deal with these objects when they are close, and transfer this
knowledge from a task to another).
\end{description}

\noindent
We consider tasks of dimensions similar to Figure \ref{fig1} and with a variable number of fires and pits.
The number of episodes for all the tasks is the same.

\subsection{Algorithms and implementation details}
We analyse different optimization techniques to solve the curriculum learning problem. In particular, we compare five different methods:
\begin{itemize}
\item \textbf{C$_0$:} where no curriculum learning is performed, i.e. $c = \emptyset$, but the agent is trained directly to solve the final task $m_L$ with starting point $\theta^L = 0^K$.
\item \textbf{GREEDY Par: } Greedy algorithm which constructs the curriculum incrementally by considering at each iteration the $n$ tasks which mostly improve the final performance \cite{curriculum2019preprint}. This is used as benchmark.
\item \textbf{GP: } where problem (\ref{eq: gray-box problem}) is modeled through a Gaussian Process and new points are drawn by maximizing an acquisition function, the Expected Improvement (EI), with a BFGS method (GPyOpt library). Since it is used without incorporating any \emph{a priori} knowledge, it searches for the best values ($\overline u, \overline p$) on the box $[0,1000]^n\times [0,100]^{n \times (n-1)}$.
\item \textbf{Heuristic} where a good estimate for $(\overline u,\overline p)$ is computed through formulas (\ref{eq: p}) and (\ref{eq: u}), with $\overline U$ calculated such that $\min_{(i,j)} \overline p_{ij} \geq 0$ and $\min_{i} \overline u_{i} \geq 10 \max_{(i,j)} \overline p_{ij}$.
\item \textbf{TPE:} where the surrogate model of problem (\ref{eq: gray-box problem}) is defined by a Tree-structured Parzen Estimator and new points are drawn by maximizing the EI (hyperopt library). It is used as a local-search method by defining the distribution of ($u,p$) as a gaussian distribution centered in the values ($\overline u, \overline p$) returned by the heuristic and with a variance proportional to the mean of ($\overline u, \overline p$) respectively.

\end{itemize}

\noindent
The proposed framework is implemented in Python 3.6 on a Intel(R) Core(TM) i7-3630QM CPU 2.4GHz. by means of the following libraries:
\begin{description}
\item[docplex (v 2.8.125):] version of Cplex used for solving the ILP (\ref{eq: scheduling}). We set the running time to 60 seconds per iteration and the mipgap to $10^{-2}$.

\item[GPyOpt (v 1.2.5):] used as black-box optimization algorithm for solving problem (\ref{eq: gray-box problem}) when no information on good estimates of $(\overline u,\overline p)$ is available. It is a Sequential Model Based Optimization (SMBO) algorithm where the surrogate function is defined through a Gaussian Process  and the new point is determined by the maximization of the EI \cite{gpyopt2016,rasmussen2004gaussian}.

\item[hyperopt (v 0.2):] used as black-box optimization algorithm for solving problem (\ref{eq: gray-box problem}) when a good estimate of  $(\overline u,\overline p)$ is available. It is an SMBO method where the surrogate function is defined by a Tree-structured Parzen Estimator and the new point is determined as in the previous case by maximizing the acquisition function \cite{hyperopt2,hyperopt,bengio}.

\item[Burlap:] used for the implementation of the GridWord domain along with the Sarsa($\lambda$) code as learning algorithm to update the policy and Tile Coding as the function approximator (\textit{http://burlap.cs.brown.edu}). 

\end{description}

\subsection{Numerical results}
We consider two different experiments on the GridWorld domain. In the first example, we define $n=12$ different tasks and we impose that at maximum $L=4$ tasks of them can be performed, obtaining 13345 potential curricula. For this example we set $N_m = 300$. In the second case, $n=7$ tasks are defined and all of them can be considered in the same curriculum $L=7$, for a total of 13700 possible combinations of tasks.
For this example we set $N_m = 400$. See \cite{curriculum2019preprint} for futher details about these examples.

Algorithm C$_0$ requires 1 curriculum evaluation, i.e. call of ${\cal P}_r$, while Heuristic needs $n^2$ curriculum evaluations.
The number of curriculum evaluations granted to the other algorithms is 300.

In Table \ref{tab: results} for each algorithm we report:
\begin{itemize}
\item the best value of the regret found (${\cal P}_r$)
\item the ranking of the returned solution with respect to all the possible curricula ({\bf rank})
\end{itemize}

% Table generated by Excel2LaTeX from sheet 'Foglio2'
\begin{table}
\centering
{\scriptsize
  \caption{Results obtained on GridWorld domain problem (${\cal P}_r^*$ indicates the regret obtained with the optimal policy).}
    \begin{tabular}{ccccccc}
    & & \multicolumn{2}{c}{n = 12, L = 4} & & \multicolumn{2}{c}{n = 7, L = 7} \bigstrut[b]\\
    \hline
    \textbf{Algorithm} &  & ${\cal P}_r$ & \textbf{rank} & & ${\cal P}_r$ & \textbf{rank} \bigstrut\\
    \hline
    C$_0$    & & -0,6389 & 11499 & & -0,5051  & 4535  \bigstrut\\
    GREEDY Par & $\qquad$ & -0,7765 & 144   & & -0,6113 & 260 \bigstrut\\
    GP    & & -0,7882 & 32    & & -0,6511 & 38 \bigstrut\\
    Heuristic & & -0,7773 & 121   & & -0,5966 & 417 \bigstrut\\
    TPE   & & -0,8025 & 4     & & -0,6697 & 14 \bigstrut\\
    \hline
    & & ${\cal P}_r^*: -0,8149$, & $|{\cal C}|=13345$ & $\qquad$ & ${\cal P}_r^*: -0,7224$, & $|{\cal C}|=13700$
    \end{tabular}%
  \label{tab: results}%
    }
\end{table}%

\noindent
From the numerical results, it is evident how all the proposed optimization methods based on the gray-box are able to improve the performance value ${\cal P}_r$ obtained when training the agent directly on the final task (algorithm C$_0$). As a proof of the effectiveness of the proposed heuristic method from (\ref{eq: p}) and (\ref{eq: u}), we highlight how this procedure is always able to find better solutions than $C_0$ and similar solutions to those returned by GREEDY Par. Moreover, the definition of a surrogate function through a Gaussian Process seems to be a successful choice in order to further improve the solution found.
Finaly, the local search performed by TPE around the tentative point $(\overline u,\overline p)$ leads to a remarkable improvement of the final performance by finding, in both the two scenarios, one of 15th best solutions out of the more than 13000 possible curricula.

\bibliographystyle{splncs04}
\bibliography{mylib}

\end{document}